\documentclass[letterpaper, 10 pt, conference]{ieeeconf}  

\IEEEoverridecommandlockouts                              
\usepackage{graphics} 
\usepackage{epsfig} 
\usepackage{times} 
\usepackage{amsmath} 
\usepackage{amssymb}  
\usepackage{color, soul}
\usepackage{graphicx}
\usepackage{caption}
\usepackage{subcaption}
\usepackage{cite}
\usepackage{changes}
\definechangesauthor[name=Ioanna]{im}
\setdeletedmarkup{\color{red} \sout{#1}}
\setaddedmarkup{\color{cyan} #1}
\DeclareMathOperator*{\argMin}{arg\,min}
\title{\LARGE \bf
	Data-Driven Model Predictive Control for the Contact-Rich Task of Food Cutting}

\author{Ioanna Mitsioni$^{1}$, Yiannis Karayiannidis$^{2}$, Johannes A. Stork$^{3}$ and Danica Kragic$^{1}$ 
	\thanks{$^{1}$ Division of Robotics, Perception and Learning (RPL), CAS, EECS, KTH Royal Institute of Technology, Stockholm, Sweden 
		{\tt\small mitsioni,dani@kth.se}}%
	\thanks{$^{2}$Division of  Systems and Control, Dept. of Electrical Engineering, Chalmers University of Technology, Gothenburg, Sweden 
		{\tt\small yiannis@chalmers.se}}%
	\thanks{$^{3}$Center for Applied Autonomous Sensor Systems (AASS), \"{O}rebro University, \"{O}rebro, Sweden. 
		{\tt\small johannesandreas.stork@oru.se}}
	\thanks{This work was supported by the Swedish Foundation for Strategic
		Research project GMT14-0082 FACT and by the Knut and Alice Wallenberg Foundation projects WASP and IPSYS.}%
}

\begin{document}

	\maketitle
	\thispagestyle{empty}
	\pagestyle{empty}

	\begin{abstract}
		Modelling of contact-rich tasks is challenging and cannot be entirely solved using classical control approaches due to the difficulty of constructing an analytic description of the contact dynamics. Additionally, in a manipulation task like food-cutting, purely learning-based methods such as Reinforcement Learning, require either a vast amount of data that is expensive to collect on a real robot, or a highly realistic simulation environment, which is currently not available. This paper presents a data-driven control approach that employs a recurrent neural network to model the dynamics for a Model Predictive Controller. We build upon earlier work limited to torque-controlled robots and redefine it for velocity controlled ones. We incorporate force/torque sensor measurements, reformulate and further extend the control problem formulation. We evaluate the performance on objects used for training, as well as on unknown objects, by means of the cutting rates achieved and demonstrate that the method can efficiently treat different cases with only one dynamic model. Finally we investigate the behavior of the system during force-critical instances of cutting and illustrate its adaptive behavior in difficult cases.
	\end{abstract}

	\section{Introduction}
	
	In contact-rich manipulation tasks, such as almost any physical interaction with objects, the contact dynamics exhibit a high degree of variation based on the type of task, as well as physical properties of the objects.
	Humans are highly compliant and able to gracefully manipulate different objects and tools by adjusting the exerted force based on their multi-modal sensory feedback, extensive training and long interaction experience.
	Depending on the task and the type of objects one interacts with, there are various parameters that affect the interaction and require different types of control.
	For example, different types of knives need to be controlled differently depending on their properties such as weight or material.
	Even when the same knife is used for cutting various types of vegetables, the controller needs to take into account the size and hardness properties of each type. 
	Furthermore, even for the same vegetable class, for example a pepper, we cannot assume that the hardness is either constant or uniform, given the various pepper types or for a single pepper, given the empty core.
	
	\begin{figure}[t]
		\centering
		\begin{subfigure}[b]{0.47\textwidth}
			\centering
			\includegraphics[width=1\linewidth]{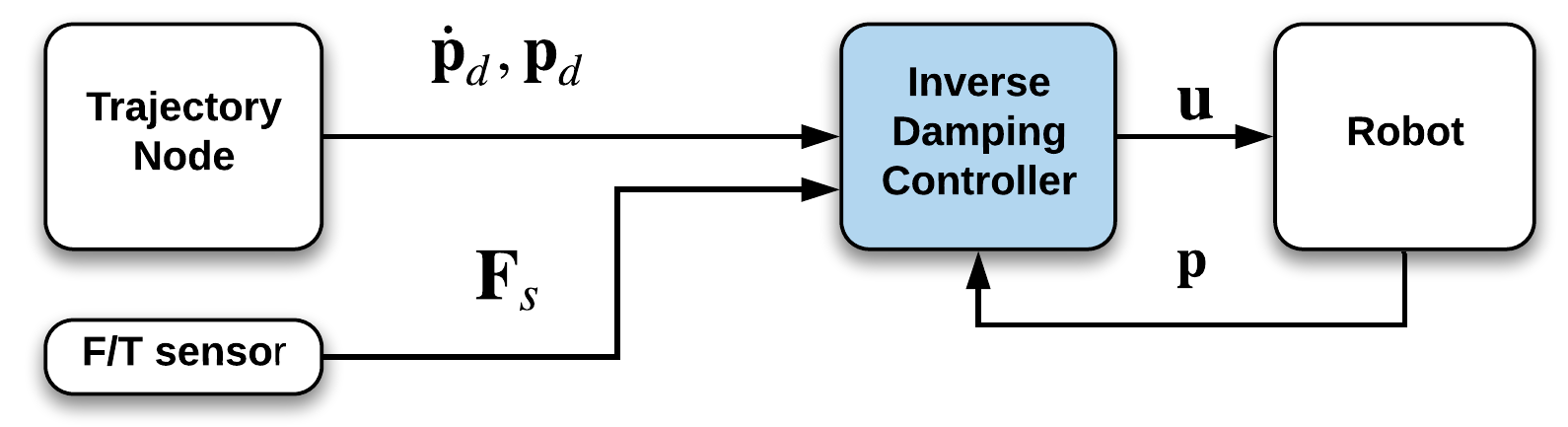}
			\caption{\footnotesize Data collection}
			\label{training}
		\end{subfigure}
		
		\begin{subfigure}[b]{.49\textwidth}
			\centering
			\includegraphics[width = 1\linewidth]{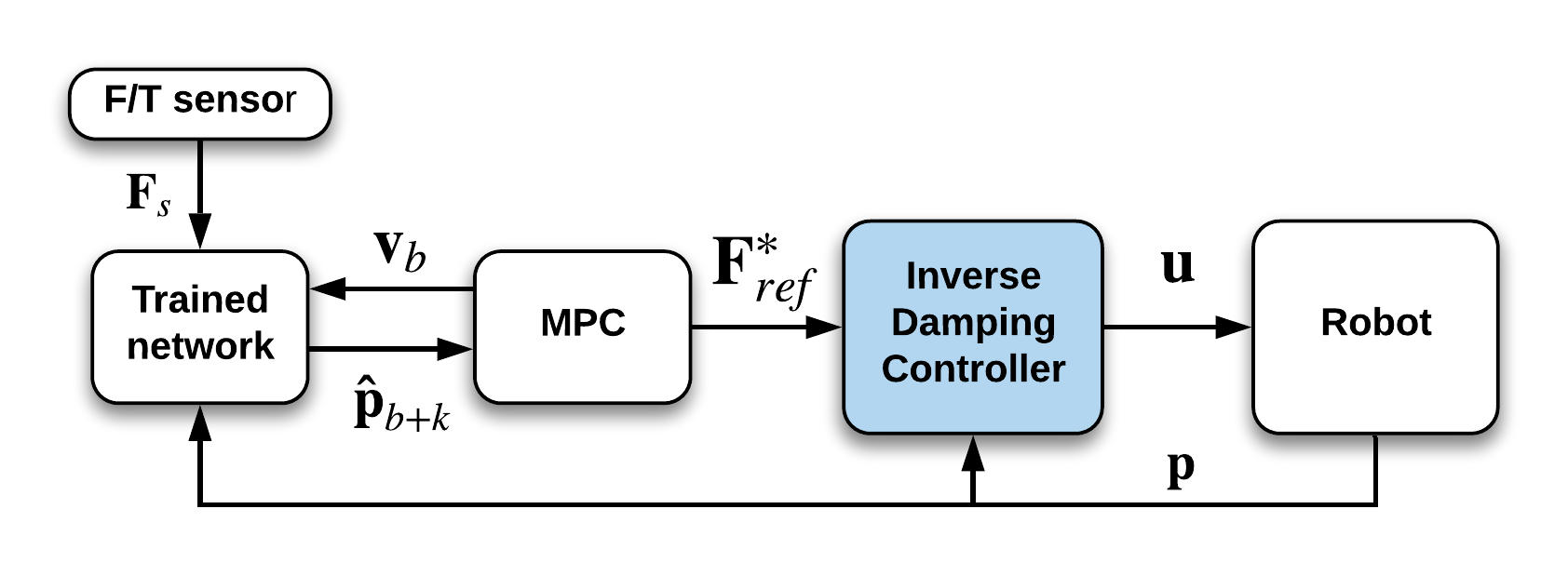}
			\caption{\footnotesize Online deployment}
			\label{online}
		\end{subfigure}
		\caption{\footnotesize System overview during data collection and online deployment}
		\label{system}
		\vspace{-0.5cm}
	\end{figure}

	In robotic manipulation, to address such parameter variations, classical force control approaches require a lot of explicit tuning.
	On the other hand, data-driven learning approaches require vast amounts of data that may be considerably expensive or impossible to acquire in a real setting. 
	Finally, all these parameters cannot be simulated in a realistic manner given the available simulators. 
	A popular control scheme that does not require a lot of fine tuning is Model Predictive Control (MPC) \cite{RHC}.
	Controllers of this type are able to optimize the control inputs by considering the long-term performance based on an accurate model of system dynamics.
	However, for tasks exhibiting a large degree of variation in the dynamics, providing such a detailed analytic model can be challenging or even practically unattainable. 
	
	In this work, we address a challenging task, that of food cutting, by combining an MPC with a data-driven approach as shown in Fig.\ref{system}. Our system employs deep learning and recurrent neural networks to act as the basis for the controller. As a result, we obtain a solution that leverages the modeling capability of deep learning and combines it with the object independent control of MPC to address the aforementioned variations. In contrast to open-loop policies, where the complete control sequence depends only on the initial state, receding horizon control schemes, such as MPC, implement a closed loop policy with the control action depending on the measured state at every timestep. This way, potential model/plant mismatches, disturbances or discontinuities in the dynamics can be accounted for at the next sampling instant, preventing the system from exhibiting unstable behaviour. Additionally, the model and the controller itself are not coupled since MPC does not require training, so if a different behaviour needs to be generated, the user can simply change the cost function parameters to accommodate for it.
	
	We chose to examine the potential of the data-driven MPC approach in the context of robotic food-cutting as its intrinsic dynamics provide an excellent test-bed. Although the problem has previously been addressed in the seminal work of Lenz et al. \cite{deepmpc}, 
	it has not been revisited with a data-driven approach. We believe that this is due to the complexity of the task itself, as well as the challenge of adjusting the original method of \cite{deepmpc} to different robotic settings considering the differences in technical aspects such as sensory information or payloads. It should be noted that most robots used in similar research topics are torque-controlled and inherently compliant, unlike the majority of widely-used robots that are either position, or velocity controlled. This adds another level of complexity as the state variables cannot include joint torques and the control laws need to be resolved either through position or velocity. With that in mind, we are taking advantage of force/torque sensors to implement a reactive controller that encodes the desired behavior while also allowing for a task-space control scheme. 
	
	The main motivation behind our work is to find a balance between classical control approaches that require tuning and an analytic description of the task on one side, and purely learning approaches that require a vast amount of data, on the other. In this work, we thus reformulate \cite{deepmpc} to velocity controlled manipulators with no access to joint torque measurements, but that are instead equipped with force sensing. We illustrate the effectiveness of our method in learning the dynamics of the manipulation task in a series of experiments and additionally demonstrate its generalization to unknown objects. Finally, we further evaluate its adaptation ability in difficult, force-critical cases where the dynamics might lead to a failed cut.
	
	\section{Related Work}
	The goal of our work is to develop a data-driven control approach that allows for online adaptation of the system's behavior in a setting with complex dynamics. Towards this aim, we learn the physical dynamics through a recurrent deep network that allows the prediction of future system states while performing online calculations for a Model Predictive Controller. Fig. \ref{system} presents schematically the control process during data collection (Fig. \ref{training}) and online deployment (Fig. \ref{online}). Below, we summarize relevant work both in terms of classical control and data-driven approaches. 
	
	Robotic manipulation has been mainly treated through force control \cite{hogan, Siciliano:2000:RFC:555628, schutter}. Despite their robustness and good performance in simple, well-defined tasks, force controllers lack the ability to discern long-term interactions. In a complicated task that displays a lot of variations in the contact dynamics, that translates to tedious tuning in order to make the controller effective \cite{tune}. Adaptive control \cite{adaptiveRev,doors} offers a mechanism for adjusting such parameters online and can account for some degree of parametric uncertainty in the system. However, it requires much simpler models and even then, results in a greedy policy as opposed to the, locally, optimal ones we address. Parallel and hybrid position/force approaches \cite{parallel, hybrid} provide efficient formulations for a plethora of tasks but are still only applicable to simpler interaction models where the geometries can be accurately defined by hand-picking the corresponding position/force controlled axes through the selection matrices. Finally, optimal control approaches such as LQR\cite{lqr} and its variants, although capable of adapting to the observed dynamics, tend to require linear or linearized models and quadratic costs that inevitable hinders the complexity they can handle.
	
	In an attempt to overcome the difficulties of classical control, several categories of data-driven approaches have emerged in manipulation. Some of them, employ human demonstrations in order to learn the force profiles needed to successfully complete complex tasks \cite{billard1, billard2, deformable, mixtureofattractors, Akgun, Huang2016}. However, in our scenario, recording a demonstration by leading the robot through the motion, would prove problematic since it would be impossible to distinguish the wrench applied by the human from the one applied by the object without having to resort in external tracking solutions as the one in \cite{peginhole}.
	Considerable progress has been made by the reinforcement learning community \cite{pmlr-v87-kalashnikov18a, Levine2} but a large body of these methods is model-free which is  limiting due to sample complexity. On the other hand, model-based methods \cite{levine1, model1, model2, model3, model4} require less samples but employ simpler approximators such as linear representations, which are are not appropriate for highly non-linear dynamics or Gaussian process that cannot handle discontinuities. In addition, reinforcement learning requires either online exploration, which would be dangerous in this setting, or training in simulation, which is not realistic as it is infeasible to accurately model the dynamics during cutting.  
	
	Similar to our work, several researchers have used neural networks to learn the dynamics model for an MPC.  In \cite{nagabandi, InformationTheoretic} however, the authors' focus on tasks with a narrow set of dynamics and the final goal is to improve an RL policy through real world trials. 
	Lastly, in \cite{DMPCLasers, dressing} the authors combine deep networks and MPC in a manner akin to ours but in the distinct fields of self-tuning optical systems and robot-assisted dressing which contains simpler contact dynamics.

	\section{Problem Formulation}\label{formulation}
	Consider a robotic manipulator equipped with force sensing. Let $\mathbf{p} \in \mathbb{R}^3 $ denote the translation part of the end-effector pose in the world frame and $\mathbf{F}_s \in \mathbb{R}^3 $ the sensor's force measurements. Let further  $\mathbf{p}_d, \dot{\mathbf{{p}}}_d$ denote a desired position and velocity of a trajectory, $\mathbf{F}_r, \mathbf{F}_d$ the reference and desired force and $\mathbf{u}$ a velocity control input. If the goal is to follow a predefined trajectory in a compliant manner, we can employ a variation of admittance control, called the inverse damping control law
	\begin{equation}\label{general_control}
	\mathbf{u} = \mathbf{K}_a(\mathbf{F}_s - \mathbf{F}_r).
	\end{equation}
	We can then define the desired compliant behavior as
	\begin{equation}\label{damping_behavior}
	\mathbf{F}_r = \mathbf{F}_d - \mathbf{K}_a^{-1}(\dot{\mathbf{p}}_d - \mathbf{K}_p \mathbf{e}_p)
	\end{equation}
	where $\mathbf{K}_p,\, \mathbf{K}_a \in \mathbb{R}^{3\times 3} $ are the stiffness and compliance gain matrices and $\mathbf{e}_p = \mathbf{p}-\mathbf{p}_d$ is the position error.
	Substituting Eq.~\eqref{damping_behavior} in Eq.~\eqref{general_control} and noting that the control input corresponds to the Cartesian velocity, results in the desired dynamic behavior
	\begin{equation}\label{admittance_law}
	\dot{\mathbf{ e}}_p + \mathbf{K}_p \mathbf{e}_p = \mathbf{K}_a \mathbf{e}_f
	\end{equation}
	where $\dot{\mathbf{e}}_p=\dot{\mathbf{p}} - \dot{\mathbf{p}}_d$ and $ \mathbf{e}_f = \mathbf{F}_s - \mathbf{F}_d \,\in \mathbb{R}^3$ are the velocity and force errors.
	
	The variations in the contact properties during a cutting task make it impossible to globally define a fixed trajectory $\mathbf{p}_d,\, \dot{\mathbf{p}}_d$ or a desired force $\mathbf{F}_d$. For instance, for objects of different size, different types of motion are required to ensure that the knife will make initial contact. Moreover, a fast motion with large velocity that is appropriate for less stiff objects, will not be applicable to stiffer ones. Nonetheless, we can produce the desired behavior by determining a reference force $\mathbf{F}_{r}$ for the axes that are involved in the cutting motion in Eq.~\eqref{admittance_law}.
	Since the geometry and the dynamics of the contact are unknown, we model them as a discrete-time dynamics function $\hat{\mathbf{p}}_{t+1} = f(\mathbf{p}_t, {\mathbf{F}_s}_t, {\mathbf{F}_{r}}_t)$ and determine an optimal reference force such that it minimizes a cost  $C(\mathbf{p}_t ,{\mathbf{F}_{r}}_t)$ over a time horizon $T$, by solving the optimization problem
	\begin{equation}\label{optimal_control}
	\begin{split}
	\mathbf{F}_{r}^* & = \argMin_\mathbf{F_r} \sum_{k=0}^{T} C\big(\hat{\mathbf{p}}_{t+k}, {\mathbf{F}_{r}}_{t+k}\big). \\
	\end{split}
	\end{equation}
	We parametrize the dynamics function $f(\mathbf{p}_t, {\mathbf{F}_s}_t, {\mathbf{F}_{r}}_t)$ as a deep recurrent network that receives the current positions, measured and reference forces, and outputs the estimated future positions. We define the model's state as the augmented state vector $\mathbf{{x}}_t = \{{\mathbf{p}}_t,{\mathbf{{F}}_{s}}_t\}$ for brevity and denote $\mathbf{v}_t = {\mathbf{F}_{r}}_t$, resulting in the formulation
	\begin{equation}\label{dynamics}
	\mathbf{\hat{p}}_{t+1} = f(\mathbf{x}_t, \mathbf{v}_t).
	\end{equation}
	Incorporating the force measurements in the augmented state vector and defining the transition function is such a way, allows this method to be used with velocity-controlled robots. However, if a torque interface is available, substituting $\mathbf{F}_s$ with the current end-effector wrench provided by the robot's joints and $\mathbf{F}_{r}$ for the future control input, will result in the formulation proposed in \cite{deepmpc}.

	\begin{figure}[t]
		\centering
		\includegraphics[width = 0.7\linewidth]{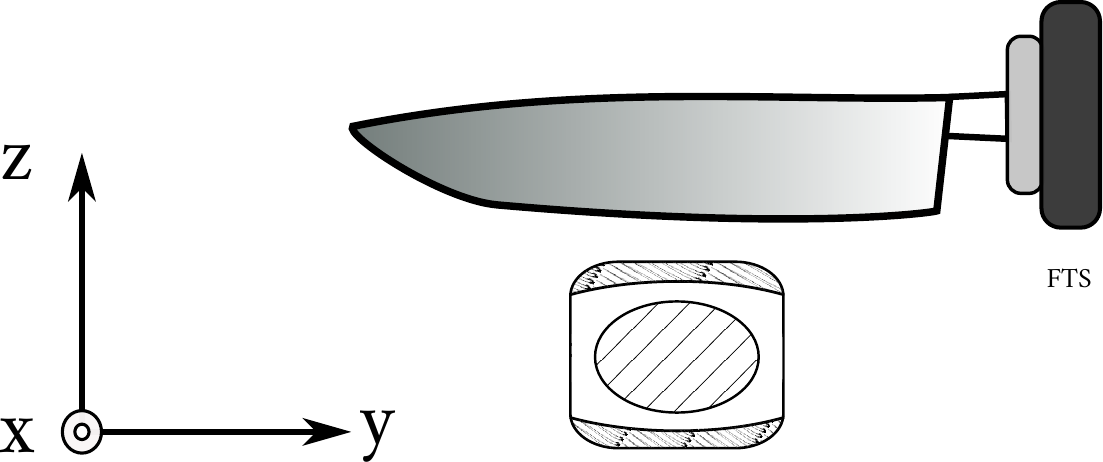}
		\caption{\footnotesize {World frame orientation. The cutting axis corresponds to Z and the sawing to Y. FTS denotes the force/torque sensors.}}
		\label{gazebo}
		\vspace{-0.3cm}
	\end{figure}
	
	We simplify the problem by considering 
	only the translational components of the motion, namely $Y$ and $Z$ as seen in Fig.\ref{gazebo}, as the state of the MPC. Since we do not require any trajectories for these axes and $\mathbf{F}_{r}^*$ acts as $\mathbf{F}_d$ during online deployment, the respective gains are set to zero. Accordingly, the rest of the axes are stabilized through a set-point stiffness control law by setting $\dot{\mathbf{x}}_d $ and the corresponding compliance gains to zero.

	\section{Method}
	In this section, we describe the architecture of the network implementing Eq.~\eqref{dynamics} and the training procedure in Sec.~\ref{net_ar}. In Sec.~\ref{data_col} we explain the data collection process and finally, in Sec.~\ref{onl_depl} we present the procedure for the online deployment of the method.
	
	\begin{figure}[t]
		\centering
		\includegraphics[width = 0.9\linewidth]{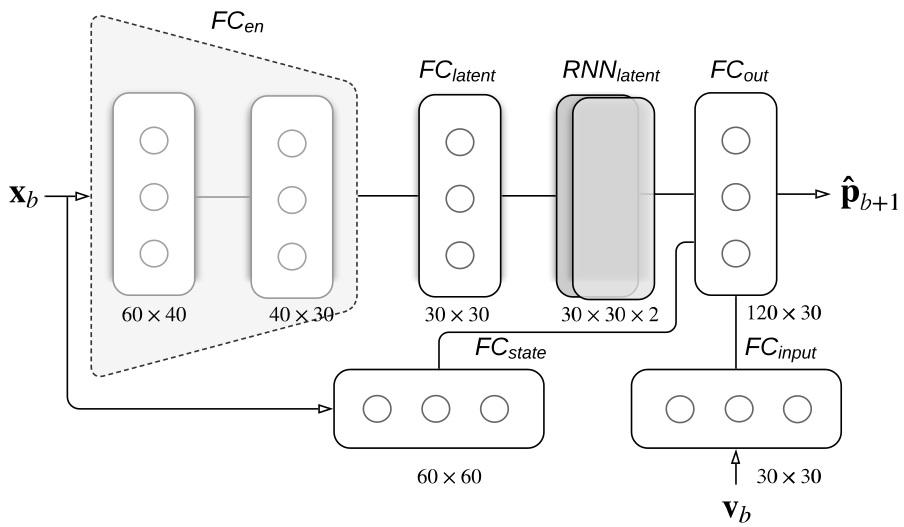}
		\caption{\footnotesize Network 
			architecture for implementing Eq.~\eqref{dynamics}.}
		\label{network}
	\end{figure}
	
	\subsection{Network Architecture and Training}\label{net_ar}
	The network architecture is depicted in Fig.~\ref{network} and consists of 6 fully connected ($FC_i$) layers and 2 recurrent layers with 30 units each ($RNN_{latent}$).
	Since the network serves as a dynamic model, it is imperative that i) the representation captures the most relevant parts of the input, and ii) that it is able to handle the task's time-varying nature. Regarding the former, in cases when contact tasks are addressed, positions and forces carry redundant information that may compromise the network's performance. By embedding the measurements in a lower-dimensional latent representation, the most salient features of the data are kept, thus allowing learning to be more effective. To accomplish this, the system state is initially processed by 2 fully connected layers that are structured as an Encoder ($FC_{en}$) and provide the latent representation of the dynamics by learning a mapping from the initial 6-dimensional input to a 3-dimensional latent feature vector.
	
	To address the time dependency of the dynamics, we employ a recurrent layer ($RNN_{latent}$) that aims at modelling long-term relationships in the latent representation before the output layer. Accordingly, the immediate dependencies on the current state and control input are treated by $FC_{state}, \: FC_{input}$ and concatenated at the output layer.
	
	Before feeding the data to the network, we applied the same preprocessing suggested by \cite{deepmpc}, namely, instead of considering single points in time, we built non-overlapping blocks consisting of $M$ timesteps of positions and forces. This allows the input vector to be encoded as a sequence that models the actual state in a more appropriate manner than a single timestep would.
	Every block $b$ then, corresponds to the sampling interval $\big[ b M, \,(b+1)M-1\big]$ leading to $\mathbf{x}_b \in \mathbb{R}^{M\times6}$ and $\mathbf{v}_b,\, \hat{\mathbf{p}}_b \in \mathbb{R}^{M\times3}$.
	The position part of $\mathbf{x}_b $ is expressed relatively to the previous block's last position. In that way, a relative displacement over time is acquired which serves as a generalized velocity. This intuitively agrees with our goal of learning interaction control as a mapping between velocities and forces by drawing a parallel to the system's mechanical impedance. Although we could have used measured Cartesian velocities immediately, in a real robotic setting it is not advisable as the measurements are usually very noisy and not easy to acquire.  Finally, both relative positions and forces in every block are normalized by removing the mean and scaled to unit variance to make sure the network weighs them equally. 
	
	Since we use relative positions that form time blocks, the sign indicates the direction of the motion. Depending on the material at hand or the tool used, the forward and backward motions might not have the same dynamics, e.g. using a serrated knife results in completely different force profiles for the two motions. To facilitate learning and ensure that negative values are represented properly, the non-linearities of the network are hyperbolic tangents that do not threshold at zero.
	
	The final network is trained to predict a sequence of $T_{horizon}$ timesteps (or $H_b = T_{horizon}/M$ time-blocks) in the future by iteratively using its previous predicted outputs as the new position inputs. During training, the corresponding force input blocks are the forces of the respective future blocks and during online deployment the actual force measurements. The MPC cost can then be calculated as
	\begin{equation}\label{recursive_cost}
	C = C(\mathbf{{p}}_{b}, \mathbf{u}_{b}) + \sum_{i=1}^{H_b}C(\hat{\mathbf{p}}_{b+i}, \mathbf{v}_{b+i}).
	\end{equation}
	To address the common problem of error accumulation in long-term predictions, we utilize the three-stage training approach that proved to be superior to random initialization according to\cite{deepmpc}. Before attempting to predict the next state, the $FC_{en}$ layer is first initialized by training it as an AutoEncoder \cite{ae} by minimizing the $L_2$ reconstruction error for the current input. During this stage, only the current state is considered and the goal is to  ensure the latent representation has captured the dynamics of the task before it is asked to predict the future positions. Next, the rest of the network, excluding the RNN layers, is trained for a single-step prediction into the future by minimizing the $L_2$ norm for the immediate next step. Finally, the weights of that model are loaded to the full network that provides the positions for the entirety of the prediction horizon. The lack of a torque interface does not allow us to record control forces during data collection. However, we assume that controller in Eq. \eqref{general_control} is capable of tracking the desired force immediately. In consequence, for training purposes  we can use the sensor forces as $\mathbf{u}_b$, in order to enable the network to directly encapsulate the effect of measured forces on the next states, instead of solely relying on their latent representation.

	\begin{figure}[!h]
		\centering
		\includegraphics[width = 0.8\linewidth]{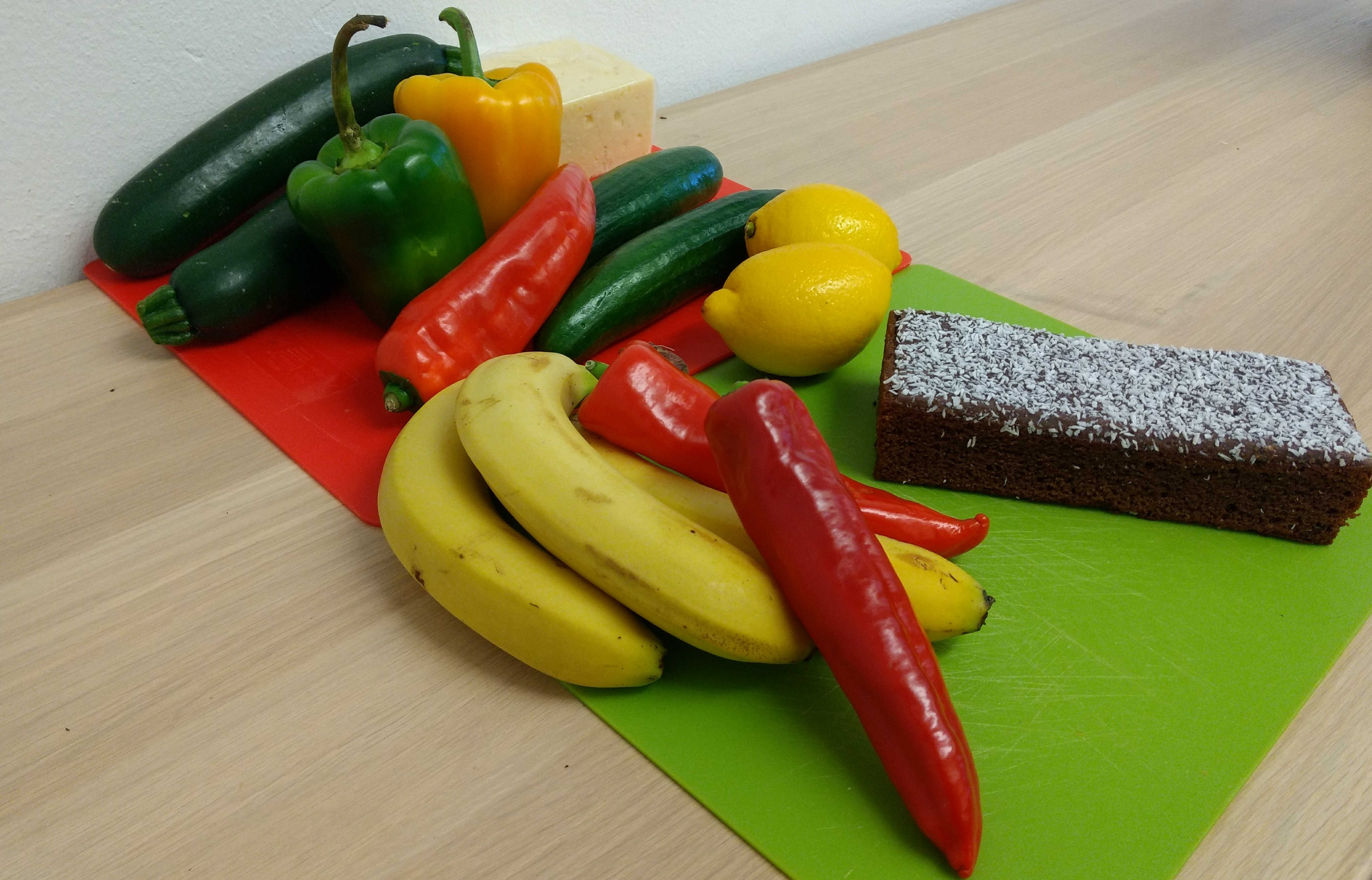}
		\caption{\footnotesize Training set consisting of fruits, vegetables and a cake. These  objects offer a wide range of interactions during cutting, spanning from homogeneous, easy to cut cases (cake, cucumber) that do not offer a lot of resistance and do not display high friction in the lateral axis of motion, to stiff objects with viscous properties (zucchinis, cheese) and finally, to objects displaying high variation in their dynamics and stiffness (bell peppers, red peppers, lemons, unpeeled bananas).}
		\label{training_set}
	\end{figure}

	\subsection{Dataset Collection}\label{data_col}
	In order to collect data, we employ the controller described in Eq.~\eqref{admittance_law} as seen in
	Fig.~\ref{training}. Our training set consisted of cutting trials on fruits, vegetables, cheese and cake, see Fig.~\ref{training_set} and had a total of 513587 data-points.
	
	For every trial, a 5th order polynomial trajectory that simulated the cutting motion was commanded as $\mathbf{p}_d, \mathbf{\dot{p}}_d$ and the stiffness and compliance gains were set manually by the operator to cover a wide range of interaction modes for every object. Force data were recorded from the sensor placed on the wrist and Cartesian positions were acquired through the robot's forward kinematics.
	The prescribed motion consisted of a periodic triangular trajectory for the sawing axis and a steady descent for the cutting one. The sawing range was kept constant for every trial but we changed the number of repetitions for thicker objects to ensure a complete cut.

	During data collection, we used clamps to stabilize the objects on the table. Although our robot is bi-manual, immobilizing the object with a set-point stiffness controller on the second arm would make that arm part of the dynamical system through the restoring forces it exerts. This would introduce additional dynamics due to the relative motion between the object and the knife, which can either enable or impede the cutting itself.


	\subsection{Online Model Predictive Control}\label{onl_depl}
	
	During online deployment, we follow the process shown in Fig.\ref{online}. In that setting, we first allow a small motion driven by the controller that was used for data collection in order to make contact with the object and initialize the system.
	
	Solving Eq.~(\ref{optimal_control}) for non-linear systems is not trivial even in the case of quadratic costs as the time the optimizer requires to converge might not allow for real-time control. Instead, we employ a shooting method \cite{nagabandi, shooting, dressing}  and demonstrate that it can achieve effective cutting rates for different object classes. For every iteration, shooting methods generate $K$ potential control inputs that act as the feasible forces for this optimization round. The solution associated with the lowest cost is then chosen as $\mathbf{F}_{r}^*$. When the state space increases, shooting methods might be slower than the optimization approach but in our case, this method is preferable as it is much simpler and allows to implicitly set input constraints by modifying the distributions that we sample inputs from.
	
	After the initial motion, for every MPC iteration, the number of feasible actions for the shooting method is generated by sampling from a uniform distribution that limits the amplitude of the generated forces to 8N. This range was chosen heuristically to be appropriate both for the robot's limits and for cutting most objects. To make the solution computationally more tractable, every time we generate an action, we assume that it will stay constant during the prediction horizon. The network is queried to predict the next states resulting from these actions, giving us the next Cartesian positions for which the accumulated cost is calculated and the best one is chosen as the next control input. The cost function used comprised two basic terms that encouraged the sawing motion around the center point $\mathbf{p}_{center}$ of the sawing range while moving downwards until the knife reached the table's height $\mathbf{p}_{table}$, a terminal cost and the norm of the control inputs, namely for every $H_b$-block horizon the cost was given by
	\begin{align}
	C(\mathbf{p}, \mathbf{v}) & = c_{\textrm{cut}}\sum_{k = 1}^{H_b}\left( \mathbf{p}^z_k - \mathbf{p}_{table}\right)^2 + \mathbf{p}^z_{H_b}\\&+ c_{\textrm{saw}}\sum_{k = 1}^{H_b}\left( \mathbf{p}^y_k - \mathbf{p}_{center}\right)^2 + c_v\sum_{k = 1}^{H_b} \left\Vert  \mathbf{v}_k \right\Vert^2 \nonumber
	\end{align}
	where $c_{\textrm{cut}}$, $c_{\textrm{saw}}$ are positive constants weighting the contribution of the costs associated with cutting and sawing actions respectively to the total cost while $c_u$ is the weighting constant for the control input quadratic term.


	\section{Evaluation}
	
	In the experiments, we used a YuMi-IRB 14000 collaborative robot manufactured by ABB with an OptoForce 6axis Force/Torque sensor mounted on its wrist. The MPC operated at a rate of 10 Hz.
	
	\begin{figure}[t]
		\centering
		\includegraphics[width=0.85\linewidth]{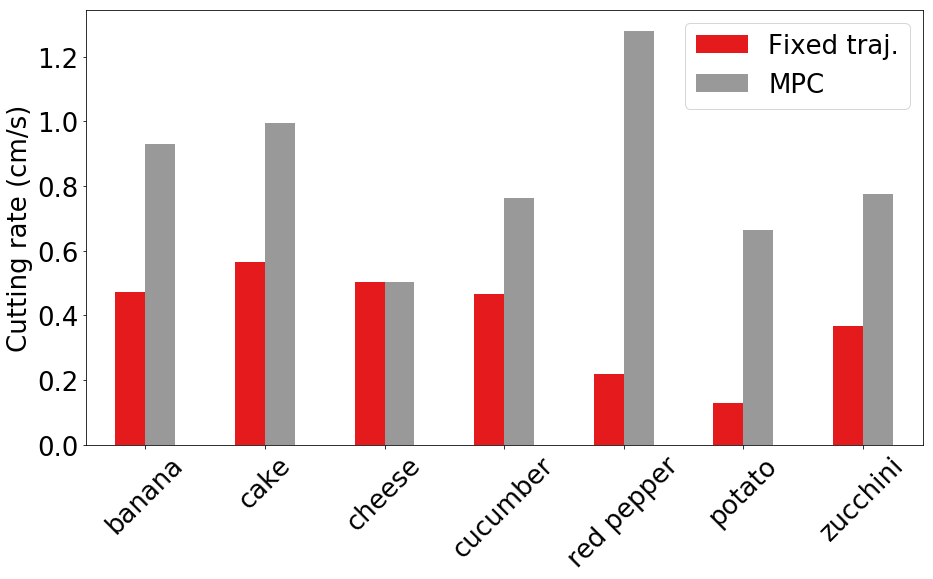}
		\caption{\footnotesize Cutting rates obtained with a fixed trajectory controller and this method}
		\label{rates}
		\vspace{-0.5cm}
	\end{figure}
	
	\subsection{Cutting Rate}\label{evaluation}
	To evaluate the proposed method, we conducted over a hundred experiments where the goal was to cut through the object completely. We deliberately included objects that offered a wide range of interactions during cutting, spanning from homogeneous easy to cut cases (cake, cucumber) that do not offer a lot of resistance and do not display high friction in the lateral axis of motion, to stiff objects with viscous properties (zucchinis, cheese) and finally, to objects displaying high variation in their dynamics and stiffness (bell peppers, red peppers, lemons, unpeeled bananas). All of these types had been encountered in training, but we additionally included the difficult case of an unseen object, i.e. potatoes, to investigate the system's behavior under unknown dynamics. For every object class we conducted 5 trials.
	
	\begin{figure}[t]
		\centering
		\begin{subfigure}[b]{0.47\textwidth}
			\centering
			\includegraphics[width=1\linewidth]{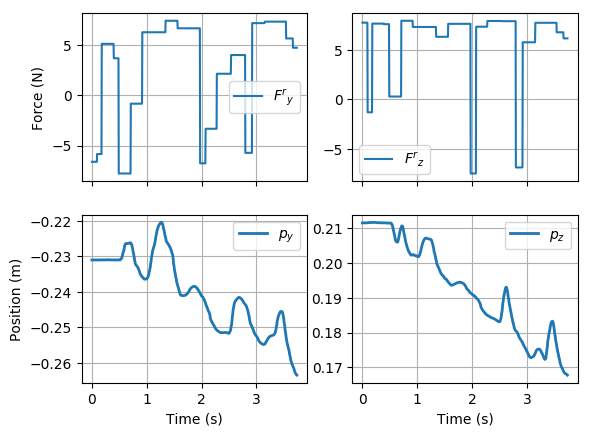}
			\vspace{-1cm}
			\caption{\footnotesize MPC}
			\label{potato_mpc}
		\end{subfigure}
		
		\begin{subfigure}[b]{.47\textwidth}
			\centering
			\includegraphics[width = 1\linewidth]{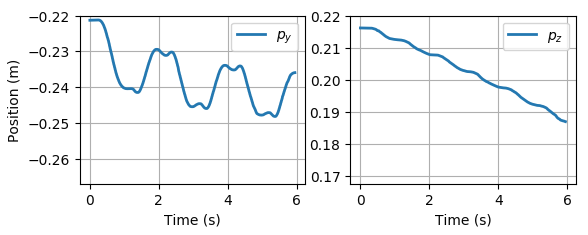}
			\vspace{-0.4cm}
			\caption{\footnotesize Fixed trajectory}
			\label{potato_adm}
		\end{subfigure}
		\caption{\footnotesize Performance while cutting a potato. The fixed trajectory controller fails to overcome friction in order to complete completely cut through the object and stop at a height of approx 0.185~cm.}
		\label{potato}
		\vspace{-0.5cm}
	\end{figure}

	\begin{figure*}[ht]
		\centering
		\includegraphics[width = 0.45 \textwidth]{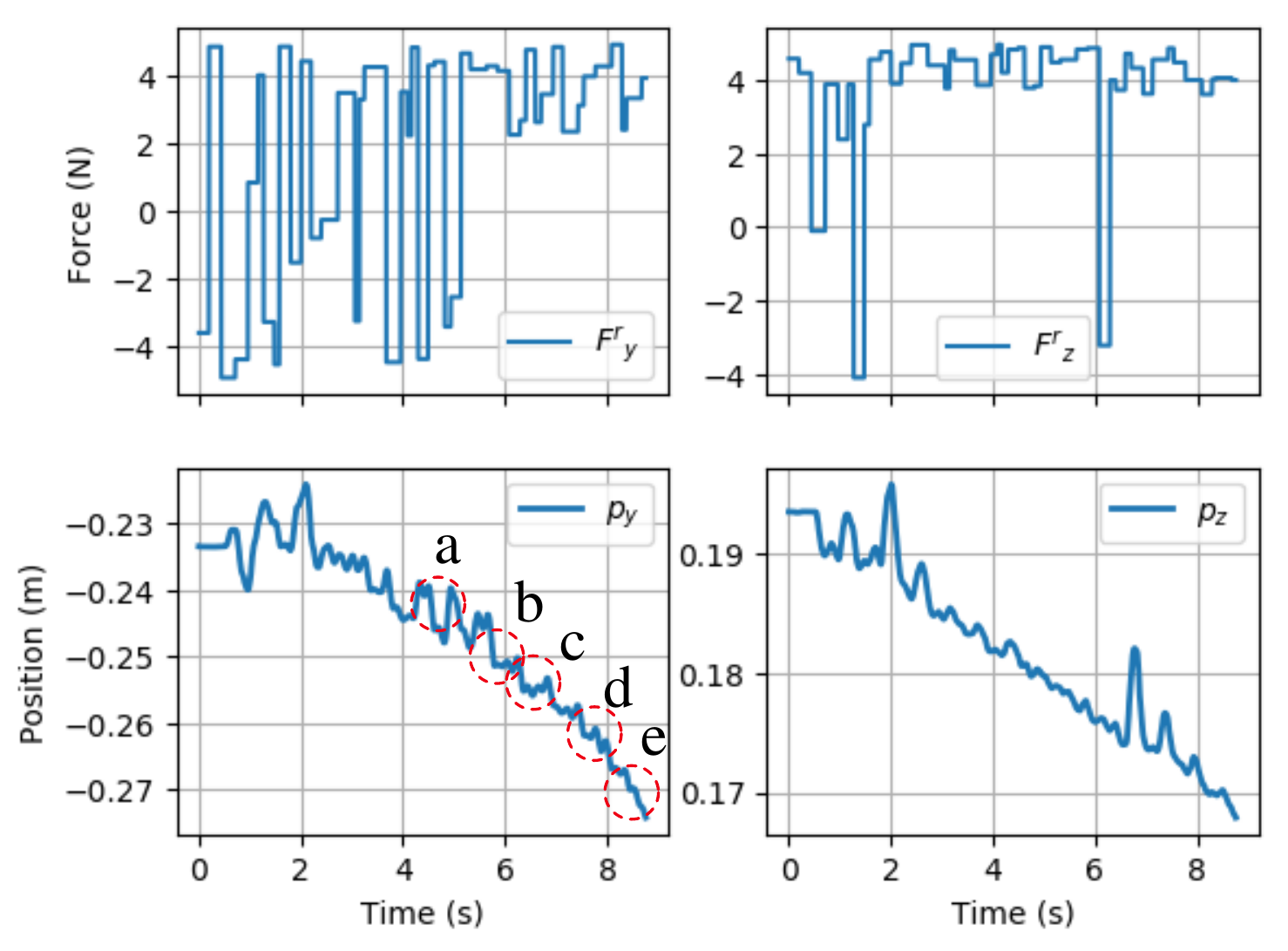}
		\label{carrots_new}
	\end{figure*}
	\begin{figure*}[h]
		\centering
		\begin{tabular}{ccccc}
			\includegraphics[width=0.17\textwidth]{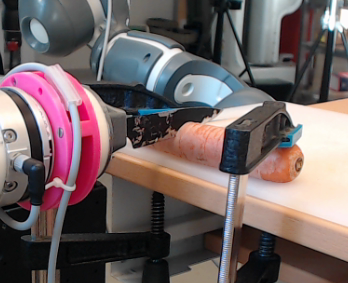}&
			\includegraphics[width=0.17\textwidth]{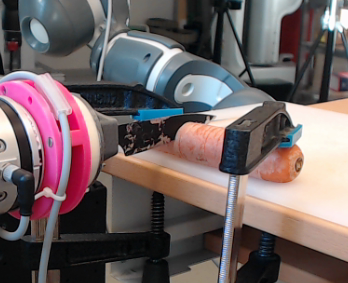}&
			\includegraphics[width=0.17\textwidth]{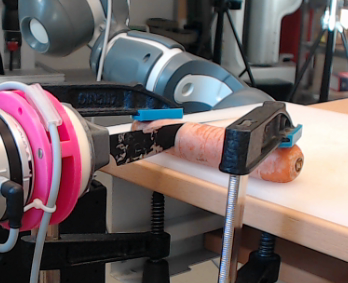}&
			\includegraphics[width=0.17\textwidth]{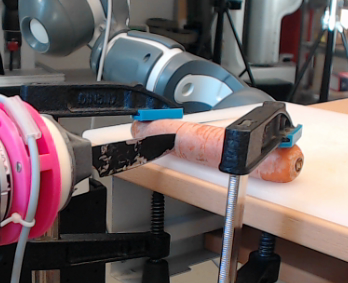}&
			\includegraphics[width=0.17\textwidth]{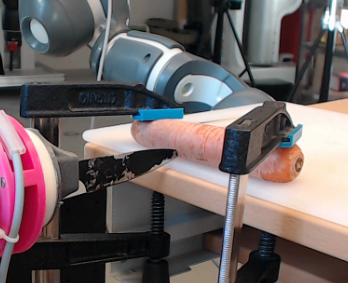}\\
			a) & b) & c) & d) & e)
		\end{tabular}
		\caption{\footnotesize Evaluation at force-critical points. The letters correspond to the respective cutting instances denoted in the figure.}
		\label{carrots} 
	\end{figure*}

	As a proof of concept, the performance was evaluated, similarly to \cite{deepmpc}, in terms of the cutting rate achieved by the proposed method as compared to the ones from the standard trajectory-tracking controller we used for data collection. To make the comparison fair, we tuned the baseline controller for each class separately by modifying both the stiffness and the trajectory followed to increase the cutting rate as much as possible for every object type.
	From the experimental results seen in Fig. \ref{rates} it is evident that our method outperforms the baseline, except in the case of cheese where the smoother motions produced by the fixed trajectory controller were more effective than their aggressive MPC counterparts to break friction. The difference in rates achieved is more palpable for the red peppers and the potatoes. In the former case, the object at hand is not homogeneous with consistent density so cutting through the skin and moving to the hollow interior directly alleviates the resistance, thus allowing for an immediate downward motion. Contrary, in the latter case, the MPC could overcome the resistance induced by the knife being fully embedded in the flesh of the potato by switching to more high-frequency sawing motions (Fig. \ref{potato_mpc}), as opposed to the fixed trajectory controller (Fig.~\ref{potato_adm}).


	\subsection{Force-Critical Points}\label{discussion}

	The ultimate goal in this work is to develop a flexible manipulation scheme that is able to adjust its behavior and respect its limitations in order to successfully complete the objective. To accomplish this, it is necessary to come up with an evaluation criterion that encompasses both of these aspects and is indicative of the behavior we are seeking. For instance, although the dynamics model's prediction accuracy is an intuitive measure to this end, a small error is not enough to reflect the network's ability to discern the dynamic behavior. Even with the normalization of the input vectors, the kinematic features have a more easily discernible structure when compared to the noisy, instant-dependent forces, which can lead the network to concentrate more on them. This can prove problematic in force-critical cases as simply moving towards the goal when progress is halted by a very stiff material, can potentially lead to failure due to joint torque limits. 
	
	Additionally, despite that cutting rate is also an intuitive measure for this task, its applicability and expressiveness are strongly correlated to the robot's potential at exerting forces and the controller used as baseline. Arguably, it is possible to design control laws that maximize the cutting rate, albeit with tuning between executions, thus confirming our first goal. However, consider the extreme case of a strong industrial manipulator. A far faster cutting rate could be achieved simply by using a position controller since the resistance from the object would never exceed the arm's potential but in our setting, this is not relevant primarily due to the hardware at hand but mostly as this criterion fails to depict the aforementioned desired behavior.
	
	To this end, we focus at force critical points where the dynamics are imperative and would lead to a failed cut. To evaluate this, we chose to use carrots as they were not encountered during training and their core is almost impossible to cut through with our robotic setting. As seen in Fig. \ref{carrots}, the normal cutting strategy fails the moment the knife hits the thicker center part of the carrot so it retreats until the tip is no longer in contact with that part. The corresponding figures in Fig. \ref{carrots} show the controller's outputs and the resulting positions during the same test. From $t = 0 \,s$ until $t = 4.5\,s$ we observe an intense sawing motion to break the friction which enables the downwards motion until approximately $t = 6\,s$ where the knife reaches the stiff part, the controller tries to lift it to re-apply pressure while only commanding positive forces on Y that pull the knife away from the force-critical point in order to minimize the cost. 
	

	\section{Conclusions and future work}\label{conclusions}
	In this article, we presented a data-driven MPC approach for the challenging task of food-cutting. We reformulated an older work by incorporating measurements from a force/torque sensor and restated the control problem. Instead of solely considering previously encountered objects during evaluation, we demonstrated that, with the same model, this method can generalize to unseen objects and treat different object types by adjusting its behavior. We investigated the system's failure modes by introducing an object that warranted an unsuccessful cut and observed the resulting trajectory. Although the outcome is a sensible solution, the direct effect that the learned dynamics model has on this behavior is unclear and will be further investigated in future work.
	
	Robotic food cutting is however only one example of numerous possible applications for this type of controllers. Integrating a complex, non-linear data-driven model into MPC, can potentially allow for elegant solutions that are able to adapt without the need for fine-tuning. This would add significant flexibility in difficult manipulation scenarios such as ones involving deformable objects where the contact dynamics are also challenging to model. 
	
	An interesting addition to our work, would be to incorporate the second arm as an explicit part of the system and explore the dynamics of the coupled motion. By actively controlling the second arm, we can induce more motion when necessary to move the knife faster out of areas with considerable friction in the lateral axis, or stabilize the object better to accelerate the motion. However, by doing so, the control task becomes even more complicated and the state space will increase accordingly. As a consequence, shooting will be insufficient and there will be need for a more sophisticated solution such as the one presented in \cite{convex}, that offers guarantees of global optimality.

	\bibliographystyle{IEEEtran}
	\bibliography{references}

\begin{thebibliography}{10}
\providecommand{\url}[1]{#1}
\csname url@samestyle\endcsname
\providecommand{\newblock}{\relax}
\providecommand{\bibinfo}[2]{#2}
\providecommand{\BIBentrySTDinterwordspacing}{\spaceskip=0pt\relax}
\providecommand{\BIBentryALTinterwordstretchfactor}{4}
\providecommand{\BIBentryALTinterwordspacing}{\spaceskip=\fontdimen2\font plus
\BIBentryALTinterwordstretchfactor\fontdimen3\font minus
  \fontdimen4\font\relax}
\providecommand{\BIBforeignlanguage}[2]{{%
\expandafter\ifx\csname l@#1\endcsname\relax
\typeout{** WARNING: IEEEtran.bst: No hyphenation pattern has been}%
\typeout{** loaded for the language `#1'. Using the pattern for}%
\typeout{** the default language instead.}%
\else
\language=\csname l@#1\endcsname
\fi
#2}}
\providecommand{\BIBdecl}{\relax}
\BIBdecl

\bibitem{RHC}
D.~Q. {Mayne} and H.~{Michalska}, ``Receding horizon control of nonlinear
  systems,'' \emph{IEEE Transactions on Automatic Control}, vol.~35, no.~7, pp.
  814--824, July 1990.

\bibitem{deepmpc}
I.~Lenz, R.~A. Knepper, and A.~Saxena, ``Deepmpc: Learning deep latent features
  for model predictive control.'' in \emph{Robotics: Science and Systems},
  2015.

\bibitem{hogan}
N.~{Hogan}, ``Impedance control: An approach to manipulation,'' in \emph{1984
  American Control Conference}, June 1984, pp. 304--313.

\bibitem{Siciliano:2000:RFC:555628}
B.~Siciliano and L.~Villani, \emph{Robot Force Control}, 1st~ed.\hskip 1em plus
  0.5em minus 0.4em\relax Norwell, MA, USA: Kluwer Academic Publishers, 2000.

\bibitem{schutter}
\BIBentryALTinterwordspacing
J.~D. Schutter and H.~V. Brussel, ``Compliant robot motion i. a formalism for
  specifying compliant motion tasks,'' \emph{The International Journal of
  Robotics Research}, vol.~7, no.~4, pp. 3--17, 1988. [Online]. Available:
  \url{https://doi.org/10.1177/027836498800700401}
\BIBentrySTDinterwordspacing

\bibitem{tune}
Y.~Karayiannidis and Z.~Doulgeri, ``An adaptive law for slope identification
  and force position regulation using motion variables,'' vol. 2006, 06 2006,
  pp. 3538 -- 3543.

\bibitem{adaptiveRev}
\BIBentryALTinterwordspacing
D.~Zhang and B.~Wei, ``{A review on model reference adaptive control of robotic
  manipulators},'' \emph{Annual Reviews in Control}, vol.~43, pp. 188--198,
  2017. [Online]. Available:
  \url{http://www.sciencedirect.com/science/article/pii/S1367578816301110}
\BIBentrySTDinterwordspacing

\bibitem{doors}
Y.~{Karayiannidis}, C.~{Smith}, F.~E.~V. {Barrientos}, P.~{\"{O}gren}, and
  D.~{Kragic}, ``An adaptive control approach for opening doors and drawers
  under uncertainties,'' \emph{IEEE Transactions on Robotics}, vol.~32, no.~1,
  pp. 161--175, Feb 2016.

\bibitem{parallel}
S.~{Chiaverini} and L.~{Sciavicco}, ``The parallel approach to force/position
  control of robotic manipulators,'' \emph{IEEE Transactions on Robotics and
  Automation}, vol.~9, no.~4, pp. 361--373, Aug 1993.

\bibitem{hybrid}
M.~H.~Raibert and J.~J.~Craig, ``Hybrid position/force control of
  manipulator,'' \emph{Journal of Dynamic Systems Measurement and Control},
  vol. 103, 12 1980.

\bibitem{lqr}
V.~Mehrmann, ``The autonomous linear quadratic control problem : Theory and
  numerical solution,'' vol. 163, 01 1991.

\bibitem{billard1}
K.~Kronander and A.~Billard, ``Learning compliant manipulation through
  kinesthetic and tactile human-robot interaction,'' \emph{Haptics, IEEE
  Transactions on}, vol.~7, pp. 367--380, 07 2014.

\bibitem{billard2}
M.~{Li}, H.~{Yin}, K.~{Tahara}, and A.~{Billard}, ``Learning object-level
  impedance control for robust grasping and dexterous manipulation,'' in
  \emph{2014 IEEE International Conference on Robotics and Automation (ICRA)},
  May 2014, pp. 6784--6791.

\bibitem{deformable}
A.~X. {Lee}, H.~{Lu}, A.~{Gupta}, S.~{Levine}, and P.~{Abbeel}, ``Learning
  force-based manipulation of deformable objects from multiple
  demonstrations,'' in \emph{2015 IEEE International Conference on Robotics and
  Automation (ICRA)}, May 2015, pp. 177--184.

\bibitem{mixtureofattractors}
S.~{Manschitz}, M.~{Gienger}, J.~{Kober}, and J.~{Peters}, ``Mixture of
  attractors: A novel movement primitive representation for learning motor
  skills from demonstrations,'' \emph{IEEE Robotics and Automation Letters},
  vol.~3, no.~2, pp. 926--933, April 2018.

\bibitem{Akgun}
\BIBentryALTinterwordspacing
B.~Akgun and A.~Thomaz, ``Simultaneously learning actions and goals from
  demonstration,'' \emph{Auton. Robots}, vol.~40, no.~2, pp. 211--227, Feb.
  2016. [Online]. Available: \url{https://doi.org/10.1007/s10514-015-9448-x}
\BIBentrySTDinterwordspacing

\bibitem{Huang2016}
\BIBentryALTinterwordspacing
B.~Huang, M.~Li, R.~L. De~Souza, J.~J. Bryson, and A.~Billard, ``A modular
  approach to learning manipulation strategies from human demonstration,''
  \emph{Autonomous Robots}, vol.~40, no.~5, pp. 903--927, Jun 2016. [Online].
  Available: \url{https://doi.org/10.1007/s10514-015-9501-9}
\BIBentrySTDinterwordspacing

\bibitem{peginhole}
T.~Tang, H.-C. Lin, and M.~Tomizuka, ``A learning-based framework for robot
  peg-hole-insertion,'' in \emph{ASME 2015 Dynamic Systems and Control
  Conference}, 10 2015, p. V002T27A002.

\bibitem{pmlr-v87-kalashnikov18a}
\BIBentryALTinterwordspacing
D.~Kalashnikov, A.~Irpan, P.~Pastor, J.~Ibarz, A.~Herzog, E.~Jang, D.~Quillen,
  E.~Holly, M.~Kalakrishnan, V.~Vanhoucke, and S.~Levine, ``Scalable deep
  reinforcement learning for vision-based robotic manipulation,'' in
  \emph{Proceedings of The 2nd Conference on Robot Learning}, ser. Proceedings
  of Machine Learning Research, A.~Billard, A.~Dragan, J.~Peters, and
  J.~Morimoto, Eds., vol.~87.\hskip 1em plus 0.5em minus 0.4em\relax PMLR,
  29--31 Oct 2018, pp. 651--673. [Online]. Available:
  \url{http://proceedings.mlr.press/v87/kalashnikov18a.html}
\BIBentrySTDinterwordspacing

\bibitem{Levine2}
\BIBentryALTinterwordspacing
S.~Levine, C.~Finn, T.~Darrell, and P.~Abbeel, ``End-to-end training of deep
  visuomotor policies,'' \emph{J. Mach. Learn. Res.}, vol.~17, no.~1, pp.
  1334--1373, Jan. 2016. [Online]. Available:
  \url{http://dl.acm.org/citation.cfm?id=2946645.2946684}
\BIBentrySTDinterwordspacing

\bibitem{levine1}
S.~Levine and V.~Koltun, ``Learning complex neural network policies with
  trajectory optimization,'' vol.~3, 06 2014.

\bibitem{model1}
J.~{Boedecker}, J.~T. {Springenberg}, J.~{Wülfing}, and M.~{Riedmiller},
  ``Approximate real-time optimal control based on sparse gaussian process
  models,'' in \emph{2014 IEEE Symposium on Adaptive Dynamic Programming and
  Reinforcement Learning (ADPRL)}, Dec 2014, pp. 1--8.

\bibitem{model2}
R.~{Lioutikov}, A.~{Paraschos}, J.~{Peters}, and G.~{Neumann}, ``Sample-based
  informationl-theoretic stochastic optimal control,'' in \emph{2014 IEEE
  International Conference on Robotics and Automation (ICRA)}, May 2014, pp.
  3896--3902.

\bibitem{model3}
\BIBentryALTinterwordspacing
S.~Levine and P.~Abbeel, ``Learning neural network policies with guided policy
  search under unknown dynamics,'' in \emph{Advances in Neural Information
  Processing Systems 27}, Z.~Ghahramani, M.~Welling, C.~Cortes, N.~D. Lawrence,
  and K.~Q. Weinberger, Eds.\hskip 1em plus 0.5em minus 0.4em\relax Curran
  Associates, Inc., 2014, pp. 1071--1079. [Online]. Available:
  \url{http://papers.nips.cc/paper/5444-learning-neural-network-policies-with-guided-policy-search-under-unknown-dynamics.pdf}
\BIBentrySTDinterwordspacing

\bibitem{model4}
M.~Deisenroth and C.~Edward~Rasmussen, ``Pilco: A model-based and
  data-efficient approach to policy search.'' 01 2011, pp. 465--472.

\bibitem{nagabandi}
A.~Nagabandi, G.~Kahn, R.~S. Fearing, and S.~Levine, ``Neural network dynamics
  for model-based deep reinforcement learning with model-free fine-tuning,''
  \emph{arXiv preprint arXiv:1708.02596}, 2017.

\bibitem{InformationTheoretic}
G.~Williams, N.~Wagener, B.~Goldfain, P.~Drews, J.~M. Rehg, B.~Boots, and E.~A.
  Theodorou, ``{Information theoretic MPC for model-based reinforcement
  learning},'' \emph{Proceedings - IEEE International Conference on Robotics
  and Automation}, pp. 1714--1721, 2017.

\bibitem{DMPCLasers}
T.~Baumeister, S.~L. Brunton, and J.~N. Kutz, ``{Deep Learning and Model
  Predictive Control for Self-Tuning Mode-Locked Lasers},'' pp. 1--9, 2017.

\bibitem{dressing}
Z.~Erickson, H.~M. Clever, G.~Turk, C.~K. Liu, and C.~C. Kemp, ``Deep haptic
  model predictive control for robot-assisted dressing,'' \emph{arXiv preprint
  arXiv:1709.09735}, 2017.

\bibitem{ae}
\BIBentryALTinterwordspacing
Y.~Bengio, P.~Lamblin, D.~Popovici, and H.~Larochelle, ``Greedy layer-wise
  training of deep networks,'' in \emph{Advances in Neural Information
  Processing Systems 19}, B.~Sch\"{o}lkopf, J.~C. Platt, and T.~Hoffman,
  Eds.\hskip 1em plus 0.5em minus 0.4em\relax MIT Press, 2007, pp. 153--160.
  [Online]. Available:
  \url{http://papers.nips.cc/paper/3048-greedy-layer-wise-training-of-deep-networks.pdf}
\BIBentrySTDinterwordspacing

\bibitem{shooting}
A.~Rao, ``A survey of numerical methods for optimal control,'' \emph{Advances
  in the Astronautical Sciences}, vol. 135, 01 2010.

\bibitem{convex}
\BIBentryALTinterwordspacing
Y.~Chen, Y.~Shi, and B.~Zhang, ``{Optimal Control Via Neural Networks: A Convex
  Approach},'' pp. 1--26, 2018. [Online]. Available:
  \url{http://arxiv.org/abs/1805.11835}
\BIBentrySTDinterwordspacing

\end{thebibliography}

\end{document}